\pdfoutput=1

\documentclass[11pt]{article}

\usepackage{EMNLP2022}

\usepackage{times}
\usepackage{latexsym}
\usepackage{multirow}
\usepackage[T1]{fontenc}

\usepackage[utf8]{inputenc}

\usepackage{microtype}

\usepackage{inconsolata}

\usepackage{xcolor}
\usepackage{colortbl}
\usepackage{amsmath}
\usepackage{algorithm,algpseudocode}
\usepackage{mathtools}

\usepackage{amssymb}
\DeclareMathSymbol{\shortminus}{\mathbin}{AMSa}{"39}

\usepackage{booktabs}
\usepackage{pifont}

\usepackage{cancel}
\newcommand{\Sref}[1]{\S\ref{#1}}
\newcommand{\Fref}[1]{Figure~\ref{#1}}
\newcommand{\Tref}[1]{Table~\ref{#1}}

\title{Token-level Sequence Labeling for Spoken Language Understanding using Compositional End-to-End Models}

\author{Siddhant Arora${}^*$ \quad Siddharth Dalmia${}^*$ \quad Brian Yan\\ \textbf{Florian Metze \quad Alan W Black \quad Shinji Watanabe} \\
Language Technologies Institute, Carnegie Mellon University, USA\\
  \texttt{\{siddhana,sdalmia\}@cs.cmu.edu} \\
}
\begin{document}
\maketitle

\begin{abstract}
End-to-end spoken language understanding (SLU) systems are gaining popularity over cascaded approaches due to their simplicity and ability to avoid error propagation. However, these systems model sequence labeling as a sequence prediction task causing a divergence from its well-established token-level tagging formulation. We build compositional end-to-end SLU systems that explicitly separate the added complexity of recognizing spoken mentions in SLU from the NLU task of sequence labeling. By relying on intermediate decoders trained for ASR, our end-to-end systems transform the input modality from speech to token-level representations that can be used in the traditional sequence labeling framework. This composition of ASR and NLU formulations in our end-to-end SLU system offers direct compatibility with pre-trained ASR and NLU systems, allows performance monitoring of individual components and enables the use of globally normalized losses like CRF, making them attractive in practical scenarios. Our models outperform both cascaded and direct end-to-end models on a labeling task of named entity recognition across SLU benchmarks.\footnote{Our code and models will be publicly available as part of the ESPnet-SLU toolkit: \url{https://github.com/espnet/espnet} and the release can be followed here: \url{ https://github.com/espnet/espnet/pull/4735} ${}^*$Equal Contribution. Siddharth is now at Google.}
\end{abstract}

\section{Introduction}
Sequence labeling (SL) is a class of natural language understanding (NLU) tasks. These systems \textit{tag} each word in a sentence to provide insights into the sentence structure and meaning~\cite{Jurafsky}. An SL system that processes unstructured text, first encodes the context and relationships of words in the sentence using an encoder and then labels each token~\cite{lample-etal-2016-neural, dozat-etal-2017-stanfords, akbik-etal-2018-contextual}. However, when dealing with spoken utterances, sequence labeling introduces an additional complexity of also \textit{recognizing} the mentions of the labels~\cite{kubala1998named,zhai2004using}. 

SL in spoken language understanding (SLU) has been approached by two schools of thought, (1) that seek to recognize the spoken words using an Automatic Speech Recognition (ASR) engine and then tag the mentions using an NLU engine in a cascaded manner~\cite{palmer-ostendorf-2001-improving,Horlock2003DiscriminativeMF,BECHET2004207}, and (2) that seek to recognize and tag the mentions directly from speech in an end-to-end (E2E) framework~\cite{ESPnet-SLU,end-to-end1}. Prior work has shown that cascaded systems suffer due to error propagation~\cite{Tran_SLU} from the ASR into the NLU engine, which can be overcome in an E2E framework. However, unlike cascaded models, E2E systems cannot utilize the vast abundance of NLU research~\cite{SLUE} as they re-define the SL problem as a complex sequence prediction problem where the sequence contains both the tags and its mentions.

Inspired by the principles of task compositionality in SL for SLU, we seek to bring both schools of thought together. Our conjecture is that we can build compositional E2E systems that first convert the spoken utterance to a sequence of token representations~\cite{dalmia-etal-2021-searchable}, which can then be used to train token-wise classification systems as per the NLU formulation. By also conditioning our token-wise classification on speech, our compositional E2E system allows recovery from errors made while creating token representations.
We instantiate our formulation on a popular SL task of named entity recognition (NER) and (1) present the efficacy of our compositional E2E NER-SLU system on benchmark SLU datasets~\cite{SLURP,SLUE} surpassing both the cascaded and direct E2E systems \Sref{sec:main_results}. 
(2) Our compositional model consists of ASR and NLU components compatible with pre-trained ASR and NER-NLU models \Sref{sec:pretrained}. 
(3) Our E2E systems exhibit transparency towards categorizing errors by enabling the evaluation of individual components of our model in isolation \Sref{sec:transparency}. 

The paper first describes the traditional SL formulation (\Sref{sec:NLU}), and discusses shortcomings in current SLU formulations (\Sref{sec:slu}). Section \Sref{sec:compositional_e2e} presents our compositional E2E model that can overcome these shortcomings. We then evaluate these approaches towards the SL task of NER (\Sref{sec:ner}).

\section{Sequence Labeling (SL)}
\label{sec:NLU}
SL systems tag each word, $w_i$, of a text sequence, $S = \{w_i \in \mathcal{V} | i=1,\dots , N\}$ of length $N$ and vocabulary $\mathcal{V}$, with a label from a label set $\mathcal{L}$, $\{ w_i \rightarrow y_i\ | y_i \in \mathcal{L}\}$. This produces a label sequence, $Y = \{ y_i \in \mathcal{L} | i=1,\ldots ,N\}$ of the same length $N$. 
Using decision theory, sequence labeling models seek to output $\hat{Y}$ from a set of all possible tag sequence $\mathcal{L}^N$,
\begin{equation}
    \hat{Y} = \underset{Y \in \mathcal{L}^N}{\operatorname{argmax}} ~P(Y|S) \label{eq:y_hat}
\end{equation}
where $P(Y|S)$ is the posterior distribution. This posterior can be modeled using various techniques like the traditional HMM \cite{HMM_article} and MEMM \cite{memm} based modeling and more recently CRF \cite{ma-hovy-2016-end} and token classification \cite{BERT} based approaches. We discuss the latter two in detail:
\paragraph{Conditional Random Field:}\citet{CRF_og} aims to directly compute the posterior of the entire label sequence $Y$ given the sentence $S$:
\begin{equation}
    P(Y|S) = \frac{e^{F(Y,S)}}{\sum_{Y' \in \mathcal{L}^N} e^{F(Y',S)}}
\end{equation}
where $F(Y,S)$ is global score of the tag sequence $Y$ given $S$. This is modeled using a linear chain CRF which computes the global score as a sum of local scores $f(.)$ for each position in $Y$ as follows
\begin{equation}
    F(Y,S) = \sum_{l=1}^{N} f(y_{l-1},y_l,S)
\end{equation}
\citet{lample-etal-2016-neural} and \citet{transformer-crf} use contextualized neural encoders like LSTMs and transformers to model context of the entire sequence $S$ for every word $w_l$. This allows for effective modeling of $f(.)$ by using encoder representations for each word as the emissions, and maintaining a separate transition score $t_{y_{l-1}\rightarrow y_l}$ to give $F(Y,S)$:
\begin{align}
    \mathbf{h}_{1:N} &= \operatorname{encoder}(w_{1:N}) \\
    t_{y_{l-1}\rightarrow y_l} &= \operatorname{transitionScores}(|\mathcal{L}|, |\mathcal{L}|)  \\
    F(Y,S) &= \sum_{l=1}^{N} ( \mathbf{h}_{l,y_l} + t_{y_{l-1}\rightarrow y_l} )
\end{align}
\paragraph{Token Classification Model:} Since the advent of strong contextual modeling using transformer based models, sequence labeling can also be treated as token classification \cite{BERT}, a simplification over MEMM estimations \cite{memm}, with the assumption that the current tag is conditionally independent to previous tag.
\begin{align}
    P(Y|S) &= \prod_{l=1}^{N} P(y_l | \mathbf{h}_l) 
\end{align}
These models are still effective as $\mathbf{h}_l$ is able to model the full context $S$ for every word $w_l$.

In cases like NER, where an entity can span multiple words, these problems are modeled using BIO tags~\cite{ramshaw1999text}, where begin (B), inside (I) tags are added for entities and an outside (O) tag for non-entity words, extending the tag set vocabulary from $\mathcal{L}$ to $\mathcal{L}' = \{ l_B \oplus l_I | l \in \mathcal{L}\} \cup \{\text{O}\}$.%
When modeled using sub-word tokens the tags can be aligned to the first sub-word token of the word and the remaining ones can be marked with a special token $\varnothing$ giving $\mathcal{L}''= \mathcal{L}' \cup \{\varnothing\}$.

\section{Sequence Labeling in SLU}
\label{sec:slu}
Sequence Labeling in SLU introduces an added complexity of recognizing mentions on top of text-based SL tasks (\Sref{sec:NLU}) as they aim to predict the tag and its mentions directly from a spoken sequence. Given a sequence of $d$ dimensional speech feature of length $T$ frames, $X = \{\mathbf{x}_t \in \mathbb{R}^d | t=1,\ldots, T\}$, these systems seek to estimate the label sequence $\hat{Y}$ from 
\begin{equation}
    \hat{Y} = \underset{Y \in \mathcal{L}^*}{\operatorname{argmax}} ~P(Y|X)
\end{equation}
where $P(Y|X)$ have been modeled as:
\paragraph{Cascaded SLU} \cite{BECHET2004207,parada2011oov,cascaded2} models $P(Y|X)$ from $P(Y|S)$ using an NLU framework (\Sref{sec:NLU}) and $P(S|X)$ using an ASR model \cite{povey2011kaldi,william_chen, graves2012sequence}, assuming conditional independence of $Y|S$ from $X$, 
\begin{align}
    P(Y|X) &= \sum_S P(Y|S,\cancel{X})P(S|X) \\
    &\approx \max_S P(Y|S) P(S|X) \\
    &\approx P(Y|\hat{S}) \max_S P(S|X) \\
    \hat{S} &= \underset{S \in \mathcal{V}*}{\operatorname{argmax}}~P(S|X)
\end{align}
Once $\hat{S}$ is estimated, $\hat{Y}$ can be estimated using Eq \ref{eq:y_hat}. Although this enables realizing $\hat{Y}$ using two well studied frameworks, the independence assumption doesn't allow recovery from errors in estimating $\hat{S}$.
\paragraph{Direct End-to-End SLU} \cite{ESPnet-SLU,SLUE,end-to-end1} systems avoid cascading errors by directly modeling $P(Y|X)$ in a single monolithic model. To achieve this while being able to recognize the spoken mentions, these systems enrich $Y$ with transcripts $S$, $Y^e = \{ y^e_i \in \mathcal{V} \cup \mathcal{L} | i = 1, \ldots ,N'\}$, where $N'$ is the length of $Y^e$. This can be modeled using an autoregressive decoder as:
\begin{equation}
    P(Y|X) = \prod_{i=1}^{N'} P(y^e_i | y^e_{1:i-1}, X) \label{eq:asr}\\
\end{equation}
However this new formulation cannot utilize the well studied sequence labeling framework \Sref{sec:NLU}. Additionally, this applies an extra burden of labeling along with alignment on the decoder and makes understanding the errors made by these systems particularly difficult. For example, Eq \ref{eq:asr} gives non-zero likelihood to a corrupt sequence with only labels and no words as $y^e \in \{\mathcal{V} \cup \mathcal{L} \}$.

\section{Compositional End-to-End SLU}
\label{sec:compositional_e2e}

\begin{figure}
    \centering
    \includegraphics[width=\linewidth]{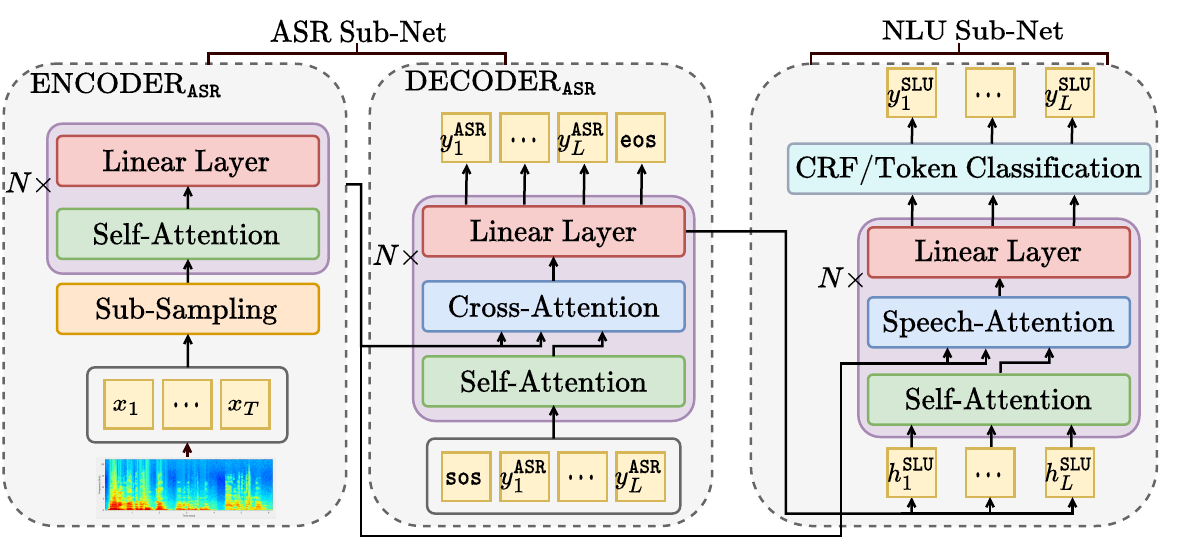}
    \caption{Schematics of our compositional E2E SLU architecture with ASR and NLU sub-nets. The ASR sub-net consists of an encoder and decoder. The NLU sub-net consists of an encoder that conditions on both speech information via $\operatorname{encoder}_\text{ASR}$ and the text information via $\operatorname{decoder}_\text{ASR}$'s hidden representation $\mathbf{h}^{\text{ASR}}$ followed by token classification or CRF layer.}
    \label{fig:model}
\end{figure}

We propose to bring the two paradigms together in a compositional end-to-end system, by extending over the cascaded SLU formulation using searchable intermediate framework \cite{dalmia-etal-2021-searchable}:
\begin{align}
    P(Y|X) &= \sum_S P(Y|S,X)P(S|X) \\
    &\approx \max_S \underset{\textsc{Sub}_\text{NLU}\textsc{Net}}{\underbrace{P(Y|S,X)}} \underset{\textsc{Sub}_\text{ASR}\textsc{Net}}{\underbrace{P(S|X)}} \label{eq:composition}
\end{align}
This system can be realized with two sub-networks as shown in Figure \ref{fig:model}, where:
\paragraph{$\textsc{Sub}_{\text{ASR}}\textsc{Net}$:} Models $P(S|X)$,
\vspace{-0.14em}
\begin{align}
&\mathbf{h}^E_{1:T} = \operatorname{encoder}_\text{ASR}(X_{1:T}) \label{comp_asr_begin}\\
&\mathbf{h}^{\text{ASR}}_{l} = \operatorname{decoder}_\text{ASR}(\mathbf{h}^E_{1:T}, w_{1:l-1}) \\
&P(w_l | X, w_{1:l-1}) = \operatorname{softmaxOut}(\mathbf{h}_l^{\text{ASR}}) \label{intermediate_eq} \\
&P(S|X) = \prod_{l=1}^{N} P(w_l | X, w_{1:l-1})
\end{align}
\paragraph{$\textsc{Sub}_{\text{NLU}}\textsc{Net}$:} Models $P(Y|S,X)$, 
\begin{align}
\mathbf{h}^{\text{NLU}}_{1:N} &= \operatorname{encoder}_\text{NLU}(\mathbf{h}_{1:N}^{\text{ASR}}, \mathbf{h}^{E}_{1:T}) \label{eq:speech_attn} \\
P(Y | S, X) &= \operatorname{CRF}(\mathbf{h}^{\text{NLU}}_{1:N}) ~~~~~~~~~~~~~~~\textsc{\textbf{OR}} \\
P(Y | S, X) &=\operatorname{TokenClassification}(\mathbf{h}^{\text{NLU}}_{1:N})\label{comp_nlu_end}  
\end{align}
The end-to-end differentiability is maintained by using $\mathbf{h}_{1:N}^{\text{ASR}}$ in Eq \ref{eq:speech_attn}. 
During inference, we approximate the Viterbi max of $S$ using beam search to give $\hat{\mathbf{h}}_{1:N}^{\text{ASR}}$. Then $\hat{Y}$ can be found using Viterbi search with no approximation as the output length is known and the solution is tractable.

This composition allows incorporating the ASR modeling and text-based sequence labeling framework \Sref{sec:NLU}. It also brings transparency to end-to-end modeling as we can also monitor performance of individual sub-nets in isolation. Further, $\operatorname{encoder}_\text{NLU}$ can attend to speech representations $\mathbf{h}^{\text{E}}_{1:T}$ using cross attention \cite{dalmia-etal-2021-searchable} enabling the direct use of speech cues for NLU. This speech attention mechanism can allow the model to recover from intermediate errors made during ASR stage.

Recently, there has been some works~\cite{raoetal2020,SaxonCMM21} that explore compositional SLU models which utilize the ASR and NLU formulations.~\citet{SaxonCMM21} uses discrete outputs from the ASR module that are made differentiable using various approaches like Gumbel-softmax \cite{jang2016categorical}. \citet{raoetal2020} also uses the ASR decoder hidden representations in the NLU module by concatenating it with token embeddings of the ASR discrete output. However, this approach requires the ASR and NLU submodule to have a shared vocabulary space, limiting the usage of pretrained ASR and LM in this architecture. Moreover, the benefits of our proposed compositional framework are not explored in these works.

\section{Spoken Named Entity Recognition}
\label{sec:ner}
To show the effectiveness of our compositional E2E SLU model we build spoken NER systems on two publically available SLU datasets, SLUE \cite{SLUE} and SLURP \cite{SLURP} (dataset and preparation details in \Sref{sec:data}). 
We compare our compositional E2E system with cascaded and direct E2E systems. 
We also compare with another compositional E2E system that  predicts the enriched transcript (\Sref{sec:slu}) using a decoder like \cite{dalmia-etal-2021-searchable} instead of label sequence (i.e. $Y^e$ instead of $Y$ in Eq.~\ref{eq:composition}) using a token level classification sub-network. We refer to this baseline model as ``Compositional E2E SLU with Direct E2E formulation''.

SLURP is evaluated using SLU-F1 \cite{SLURP} which weighs the entity labels with the word and character error rate of the predicted mentions and SLUE using F1 \cite{SLUE} which evaluates getting both the mention and the entity label exactly right. We also compute Label-F1 for both datasets which considers only the entity label. We report micro-averaged F1 for all results. %

\subsection{Model Configurations}
We build all our systems using ESPnet-SLU \cite{ESPnet-SLU} which is an open-source SLU toolkit built on ESPnet \cite{espnet}, a flagship toolkit for speech processing. We use encoder-decoder based architecture for our baseline E2E system. We use Conformer encoder blocks \cite{conformer} and Transformer decoder blocks \cite{vasvani} with CTC multi-tasking \cite{ESPnet-SLU}. The baseline compositional model with Direct E2E SLU formulation consists of a conformer encoder and transformer decoder in it's ASR component and transformer encoder and transformer decoder in it's NLU component. Our proposed compositional model with the NLU formulation, as shown in \Fref{fig:model}, replaces the NLU component in Direct E2E formulation with a transformer encoder followed by a linear layer. For the cascaded systems, we build systems that have the same size as that of our ASR and NLU sub-networks. All models were tuned separately using validation sets with the same hyperparameter search space. Full descriptions of model and training parameters are in \Sref{sec:exp_setup}.

\subsection{Performance of Compositional E2E SLU}
\label{sec:main_results}
Table~\ref{tab:main-results} shows that our proposed compositional E2E models with the token-level NLU formulation outperform both cascaded and direct E2E models on all benchmarks using both CRF and Token Classification. In order to understand gains of our proposed model, we examine the performance of our compositional system with direct E2E formulation (\Sref{sec:slu}). While being comparable to direct E2E models, they still lag behind our proposed models showing the efficacy of modeling SL tasks as a token-level tagging (\Sref{sec:NLU}) in an E2E SLU framework. 

We further analyze our compositional systems that don't attend to speech representations. We observe a performance drop as these models are not able to recover from errors made while ``recognising'' entity mentions. For example, in an utterance that says ``change the bedroom lights to green'', though the ASR component incorrectly predicts the transcript as ``change the color of lights to green'', the NLU component w/ Speech Attention is able to recover the entity type $\textsc{House\_Place}$.%
\begin{table}[t]
\resizebox {\linewidth} {!} {
  \centering
\begin{tabular}{lcc|cc}
\toprule
& \multicolumn{2}{c|}{SLURP} & \multicolumn{2}{c}{SLUE} \\ 
\cmidrule(r){2-5}
Model & SLU F1 & Label F1 & F1 & Label F1 \\ 
 \midrule
Direct E2E SLU (\citeauthor{ESPnet-SLU}) & 71.9 & - & 54.7 & 67.6 \\ %
 \midrule
Casacaded SLU (Ours) & 73.3 & 80.9 & 48.6 & 63.9  \\
Direct E2E SLU (Ours) & 77.1 & 84.0 & 54.7 & 67.6 \\  %
\midrule
Compositional E2E SLU \\
w/ Direct E2E formulation (\Sref{sec:slu}) & 77.2 & 84.6 & 50.0 & 68.0\\ %
w/ Proposed NLU formulation (\Sref{sec:compositional_e2e}) \\ 
\hphantom{000}CRF w/ Speech Attention (SA) & 77.7 & 85.2  & 59.4 & 73.6\\ %
\hphantom{000}Token Classification w/ SA & \textbf{78.0} & \textbf{85.3} & \textbf{60.3} & \textbf{73.7}\\  %
\hphantom{000}\hphantom{000}w/o Speech Attention & 77.7 & 84.9 & 59.0 & 73.6 \\  %
\bottomrule
\end{tabular}
}
\caption{Results presenting the micro F1 performance of our proposed compositional E2E models using CRF and Token Classification modeling. Cascaded, direct E2E and our compositional E2E with direct E2E formulation are shown for comparison. We also provide an ablation of our model with and without Speech Attention (SA).} 
\label{tab:main-results}
\vspace{-0.8em}
\end{table}

\subsection{Utilizing External Sub-Net models}
\label{sec:pretrained}
\begin{table}[t]
  \centering
\resizebox {\linewidth} {!} {
\begin{tabular}{lcc|cc}
\toprule
& \multicolumn{2}{c|}{SLURP} & \multicolumn{2}{c}{SLUE} \\ 
\cmidrule(r){2-5}
Model & SLU F1 & Label F1 & F1 & Label F1 \\ 
 \midrule
Direct E2E SLU & 77.1 & 84.0 & 54.7 & 67.6 \\ 
\hphantom{0}w/ NLU fine-tuning  & \multicolumn{4}{c}{Incompatible} \\
\hphantom{0}w/ ASR fine-tuning & 73.5 & 81.2 & 64.0 & 80.6\\
\midrule
Compositional E2E SLU (w/ SA) & 78.0 & 85.3 & 60.3 & 73.7\\ 
\hphantom{0}w/ NLU finetuning (w/o SA) & 77.7 & 84.9 & 62.4 & 76.4 \\
\hphantom{0}w/ ASR finetuning (w/ SA) & \textbf{81.4} & \textbf{88.8} & \textbf{71.6} & \textbf{85.2} \\
\midrule
Compositional E2E SLU (w/ SA) & 78.0 & 85.3 & 60.3 & 73.7\\ 
\hphantom{0}w/ External ASR Transcripts ($S^{\text{ext}}$) & \textbf{81.0} & \textbf{88.1} & \textbf{70.1} & \textbf{81.2} \\
\bottomrule
\end{tabular}
}
\caption{Results presenting the compatibility of our models with pre-trained ASR and NLU systems by (1) finetuning pre-trained components and (2) directly utilizing transcripts from an external ASR model.} 
\label{tab:pre-trained}
\vspace{-1.0em}
\end{table}
Components of our compositional E2E SLU model have functions similar to an ASR and NLU model (Eq \ref{comp_asr_begin}-\ref{comp_nlu_end}). This allows fine-tuning our models using sub-systems, pre-trained on large amounts of available sub-task data. \Tref{tab:pre-trained} shows that our compositional model has better compatibility with ASR and NLU fine-tuning over direct E2E systems, thereby increasing their performance gap, particularly for SLUE, an under-resourced SLU dataset.

Further our models have the ability to use transcripts from a strong external model ($S^\text{ext}$) directly during inference, by instantiating our models with these transcripts to produce $\mathbf{h}^{\text{ASR}}$ and then evaluate $P(Y|S^\text{ext},X)$. \Tref{tab:pre-trained} shows using transcripts from an external ASR \textit{with no fine-tuning steps} can achieve similar performance to ASR fine-tuning. 

\begin{table}[t]
\resizebox {\linewidth} {!} {
  \centering
\begin{tabular}{lcc|cc}
\toprule
& \multicolumn{2}{c|}{SLURP} & \multicolumn{2}{c}{SLUE} \\ 
\cmidrule(r){2-5}
& ASR & NLU & ASR & NLU \\ 
& (\%WER $\downarrow$) & (SLU-F1 $\uparrow$) & (\%WER $\downarrow$) & (F1 $\uparrow$) \\ 
 \midrule
Pure ASR \& NLU models & 16.1 & 82.4 & 30.4 & 58.1\\
\midrule
Compostional E2E SLU \\
\hphantom{0}CRF w/ Speech Attention (SA) & 16.3 & 88.3 & 27.4 & 75.6 \\ 
\hphantom{0}Token Classification w/o SA & 16.0 & 87.9 & 27.6 & 74.1 \\ 
\hphantom{0}Token Classification w/ SA & 16.1 & 88.7 & 27.5 & 75.6 \\ 
\bottomrule
\end{tabular}
}
\caption{Results showcasing the transparency of our compositional E2E models by evaluating the individual sub-networks ASR (\%WER) and NLU (F1) in isolation.}
\label{tab:sub-network-results}
\vspace{-1.0em}
\end{table}

\subsection{Transparency in Compositional E2E SLU}
\label{sec:transparency}

Following Eq \ref{eq:composition}, we can estimate ASR performance by calculating $\hat{S}$ using beam search and NLU performance by estimating $\hat{Y}$ from $P(Y | S^\textsc{gt}, X)$, where $S^\textsc{gt}$ is the ground truth transcripts. \Tref{tab:sub-network-results} shows the performances of individual components of our model along with performances of ASR and NLU only models suggesting that we can effectively monitor the performance of these components, helping practitioners analyze and debug them. For instance, while our models with and without speech attention have comparable performance on ASR, using speech attention improves NLU power. Further the one-to-one alignment of transcripts and sequence labels can provide further categorization of errors, as shown in \Sref{sec:error_cat}.

\subsection{CRF vs Token Classification}
\label{sec:CRF}
For practical SLU the likelihoods of our compositional model $P(Y|S,X)$, should be correlated with errors in label sequence $Y$. We found that in SLURP our compositional E2E SLU, while using locally normalized token classification shows no correlation (Corr=0.13,p=0), using CRF exhibits moderate correlation (Corr=0.43, p=0). This makes globally normalized models attractive for real-world scenarios like automated data auditing and human in-the-loop ML~\cite{NELL} despite their marginal addition in computation cost.

\section{Conclusion}
We propose to combine text based sequence labeling framework into the speech recognition framework to build a compositional end-to-end model for SLU. Our compositional E2E models not only show superior performance over cascaded and direct end-to-end SLU systems, but also bring the power of both these systems in a single framework.
These models can utilise pretrained sub task components and exhibit transparency like cascaded systems, while avoiding error propagation like direct end-to-end systems. 

\section*{Limitations}
Our compositional model relies on the availability of transcripts for training. This although is a limitation, it is a safe assumption for sequence labeling tasks for spoken language understanding. We can see from \Sref{sec:slu} that the task for sequence labeling in SLU also requires the model to recognize the words being spoken along with the sequence labels, implying the need for at least a partial transcript for training direct end-to-end SLU systems. 

\section*{Broader Impact}
With our compositional end-to-end SLU model, we strive to bring the research from the text based sequence labeling directly into speech based spoken language understanding. Our aim is to avoid re-invention of the wheel, but rather come up with innovative ways to build end-to-end models by converting a complex problem into simpler ones that have seen substantial research in the past. Additionally we believe the increased capacity for error analysis in our compositional end-to-end system can help towards building better practical systems during deployment. Our compositional end-to-end systems can effectively utilize pre-trained ASR and NLU systems, thereby avoiding the need for collecting large labeled datasets for SLU. This framework also saves compute by utilizing pre-trained ASR systems directly during inference to improve downstream performances with no fine-tuning.

\section*{Acknowledgement}
We thank Aakanksha Naik and the anonymous reviewers for their feedback. This work used the Extreme Science and Engineering Discovery Environment (XSEDE) \cite{xsede}, which is supported by NSF grant number ACI-1548562. Specifically, it used the Bridges system \cite{nystrom2015bridges}, which is supported by NSF award number ACI-1445606, at the Pittsburgh Supercomputing Center (PSC).

\bibliography{anthology,custom}

\begin{thebibliography}{46}
\expandafter\ifx\csname natexlab\endcsname\relax\def\natexlab#1{#1}\fi

\bibitem[{Akbik et~al.(2018)Akbik, Blythe, and
  Vollgraf}]{akbik-etal-2018-contextual}
Alan Akbik, Duncan Blythe, and Roland Vollgraf. 2018.
\newblock \href {https://aclanthology.org/C18-1139} {Contextual string
  embeddings for sequence labeling}.
\newblock In \emph{Proceedings of the 27th International Conference on
  Computational Linguistics}, pages 1638--1649, Santa Fe, New Mexico, USA.
  Association for Computational Linguistics.

\bibitem[{Arora et~al.(2022)Arora, Dalmia, Denisov, Chang, Ueda, Peng, Zhang,
  Kumar, Ganesan, Yan, Vu, Black, and Watanabe}]{ESPnet-SLU}
Siddhant Arora, Siddharth Dalmia, Pavel Denisov, Xuankai Chang, Yushi Ueda,
  Yifan Peng, Yuekai Zhang, Sujay Kumar, Karthik Ganesan, Brian Yan, Ngoc~Thang
  Vu, Alan~W. Black, and Shinji Watanabe. 2022.
\newblock \href {https://doi.org/10.1109/ICASSP43922.2022.9747674}
  {{ESPnet-SLU}: Advancing spoken language understanding through espnet}.
\newblock In \emph{{IEEE} International Conference on Acoustics, Speech and
  Signal Processing, {ICASSP} 2022, Virtual and Singapore, 23-27 May 2022},
  pages 7167--7171. {IEEE}.

\bibitem[{Bastianelli et~al.(2020)Bastianelli, Vanzo, Swietojanski, and
  Rieser}]{SLURP}
Emanuele Bastianelli, Andrea Vanzo, Pawel Swietojanski, and Verena Rieser.
  2020.
\newblock \href {https://doi.org/10.18653/v1/2020.emnlp-main.588} {{SLURP:} {A}
  spoken language understanding resource package}.
\newblock In \emph{Proceedings of the 2020 Conference on Empirical Methods in
  Natural Language Processing, {EMNLP} 2020, Online, November 16-20, 2020}.
  Association for Computational Linguistics.

\bibitem[{B{\'{e}}chet et~al.(2004)B{\'{e}}chet, Gorin, Wright, and
  Hakkani{-}T{\"{u}}r}]{BECHET2004207}
Fr{\'{e}}d{\'{e}}ric B{\'{e}}chet, Allen~L. Gorin, Jeremy~H. Wright, and Dilek
  Hakkani{-}T{\"{u}}r. 2004.
\newblock \href {https://doi.org/10.1016/j.specom.2003.07.003} {Detecting and
  extracting named entities from spontaneous speech in a mixed-initiative
  spoken dialogue context: How may {I} help you?\({}^{\mbox{sm, tm}}\)}.
\newblock \emph{Speech Commun.}, 42(2):207--225.

\bibitem[{Chan et~al.(2016)Chan, Jaitly, Le, and Vinyals}]{william_chen}
William Chan, Navdeep Jaitly, Quoc Le, and Oriol Vinyals. 2016.
\newblock \href {https://doi.org/10.1109/ICASSP.2016.7472621} {Listen, attend
  and spell: A neural network for large vocabulary conversational speech
  recognition}.
\newblock In \emph{2016 IEEE International Conference on Acoustics, Speech and
  Signal Processing (ICASSP)}, pages 4960--4964.

\bibitem[{Chen et~al.(2021{\natexlab{a}})Chen, Chai, Wang, Du, Zhang, Weng, Su,
  Povey, Trmal, Zhang, Jin, Khudanpur, Watanabe, Zhao, Zou, Li, Yao, Wang, You,
  and Yan}]{gigaspeech}
Guoguo Chen, Shuzhou Chai, Guan{-}Bo Wang, Jiayu Du, Wei{-}Qiang Zhang, Chao
  Weng, Dan Su, Daniel Povey, Jan Trmal, Junbo Zhang, Mingjie Jin, Sanjeev
  Khudanpur, Shinji Watanabe, Shuaijiang Zhao, Wei Zou, Xiangang Li, Xuchen
  Yao, Yongqing Wang, Zhao You, and Zhiyong Yan. 2021{\natexlab{a}}.
\newblock \href {https://doi.org/10.21437/Interspeech.2021-1965} {Gigaspeech:
  An evolving, multi-domain {ASR} corpus with 10, 000 hours of transcribed
  audio}.
\newblock In \emph{Interspeech 2021, 22nd Annual Conference of the
  International Speech Communication Association, Brno, Czechia, 30 August - 3
  September 2021}, pages 3670--3674. {ISCA}.

\bibitem[{Chen et~al.(2021{\natexlab{b}})Chen, Wang, Chen, Wu, Liu, Chen, Li,
  Kanda, Yoshioka, Xiao, Wu, Zhou, Ren, Qian, Qian, Wu, Zeng, and Wei}]{WavLM}
Sanyuan Chen, Chengyi Wang, Zhengyang Chen, Yu~Wu, Shujie Liu, Zhuo Chen, Jinyu
  Li, Naoyuki Kanda, Takuya Yoshioka, Xiong Xiao, Jian Wu, Long Zhou, Shuo Ren,
  Yanmin Qian, Yao Qian, Jian Wu, Michael Zeng, and Furu Wei.
  2021{\natexlab{b}}.
\newblock \href {http://arxiv.org/abs/2110.13900} {{WavLM}: Large-scale
  self-supervised pre-training for full stack speech processing}.
\newblock \emph{CoRR}, abs/2110.13900.

\bibitem[{Clark et~al.(2022)Clark, Garrette, Turc, and Wieting}]{canine}
Jonathan~H. Clark, Dan Garrette, Iulia Turc, and John Wieting. 2022.
\newblock \href {https://doi.org/10.1162/tacl_a_00448} {Canine: Pre-training an
  efficient tokenization-free encoder for language representation}.
\newblock \emph{Transactions of the Association for Computational Linguistics},
  10:73--91.

\bibitem[{Coucke et~al.(2018)Coucke, Saade, Ball, Bluche, Caulier, Leroy,
  Doumouro, Gisselbrecht, Caltagirone, Lavril, Primet, and
  Dureau}]{snips-voice-platform}
Alice Coucke, Alaa Saade, Adrien Ball, Th{\'{e}}odore Bluche, Alexandre
  Caulier, David Leroy, Cl{\'{e}}ment Doumouro, Thibault Gisselbrecht,
  Francesco Caltagirone, Thibaut Lavril, Ma{\"{e}}l Primet, and Joseph Dureau.
  2018.
\newblock \href {http://arxiv.org/abs/1805.10190} {Snips voice platform: an
  embedded spoken language understanding system for private-by-design voice
  interfaces}.
\newblock \emph{CoRR}, abs/1805.10190.

\bibitem[{Dalmia et~al.(2021)Dalmia, Yan, Raunak, Metze, and
  Watanabe}]{dalmia-etal-2021-searchable}
Siddharth Dalmia, Brian Yan, Vikas Raunak, Florian Metze, and Shinji Watanabe.
  2021.
\newblock \href {https://doi.org/10.18653/v1/2021.naacl-main.151} {Searchable
  hidden intermediates for end-to-end models of decomposable sequence tasks}.
\newblock In \emph{Proceedings of the 2021 Conference of the North American
  Chapter of the Association for Computational Linguistics: Human Language
  Technologies}, pages 1882--1896, Online. Association for Computational
  Linguistics.

\bibitem[{{Del Rio} et~al.(2021){Del Rio}, Delworth, Westerman, Huang,
  Bhandari, Palakapilly, McNamara, Dong, Żelasko, and Jetté}]{earnings21}
Miguel {Del Rio}, Natalie Delworth, Ryan Westerman, Michelle Huang, Nishchal
  Bhandari, Joseph Palakapilly, Quinten McNamara, Joshua Dong, Piotr Żelasko,
  and Miguel Jetté. 2021.
\newblock \href {https://doi.org/10.21437/Interspeech.2021-1915} {{Earnings-21:
  A Practical Benchmark for ASR in the Wild}}.
\newblock In \emph{Proc. Interspeech 2021}.

\bibitem[{Devlin et~al.(2019)Devlin, Chang, Lee, and Toutanova}]{BERT}
Jacob Devlin, Ming{-}Wei Chang, Kenton Lee, and Kristina Toutanova. 2019.
\newblock \href {https://doi.org/10.18653/v1/n19-1423} {{BERT:} pre-training of
  deep bidirectional transformers for language understanding}.
\newblock In \emph{Proceedings of the 2019 Conference of the North American
  Chapter of the Association for Computational Linguistics: Human Language
  Technologies, {NAACL-HLT} 2019, Minneapolis, MN, USA, June 2-7, 2019, Volume
  1 (Long and Short Papers)}, pages 4171--4186. Association for Computational
  Linguistics.

\bibitem[{Dozat et~al.(2017)Dozat, Qi, and Manning}]{dozat-etal-2017-stanfords}
Timothy Dozat, Peng Qi, and Christopher~D. Manning. 2017.
\newblock \href {https://doi.org/10.18653/v1/K17-3002} {{S}tanford{'}s
  graph-based neural dependency parser at the {C}o{NLL} 2017 shared task}.
\newblock In \emph{Proceedings of the {C}o{NLL} 2017 Shared Task: Multilingual
  Parsing from Raw Text to Universal Dependencies}, pages 20--30, Vancouver,
  Canada. Association for Computational Linguistics.

\bibitem[{Ghannay et~al.(2018)Ghannay, Caubrière, Estève, Camelin, Simonnet,
  Laurent, and Morin}]{end-to-end1}
S.~Ghannay, A.~Caubrière, Y.~Estève, N.~Camelin, E.~Simonnet, A.~Laurent, and
  E.~Morin. 2018.
\newblock \href {https://doi.org/10.1109/SLT.2018.8639513} {End-to-end named
  entity and semantic concept extraction from speech}.
\newblock In \emph{2018 IEEE Spoken Language Technology Workshop (SLT)}, pages
  692--699.

\bibitem[{Graves(2012)}]{graves2012sequence}
Alex Graves. 2012.
\newblock \href {http://arxiv.org/abs/1211.3711} {Sequence transduction with
  recurrent neural networks}.
\newblock \emph{CoRR}, abs/1211.3711.

\bibitem[{Gulati et~al.(2020)Gulati, Qin, Chiu, Parmar, Zhang, Yu, Han, Wang,
  Zhang, Wu, and Pang}]{conformer}
Anmol Gulati, James Qin, Chung{-}Cheng Chiu, Niki Parmar, Yu~Zhang, Jiahui Yu,
  Wei Han, Shibo Wang, Zhengdong Zhang, Yonghui Wu, and Ruoming Pang. 2020.
\newblock \href {https://doi.org/10.21437/Interspeech.2020-3015} {Conformer:
  Convolution-augmented transformer for speech recognition}.
\newblock In \emph{Interspeech 2020, 21st Annual Conference of the
  International Speech Communication Association, Virtual Event, Shanghai,
  China, 25-29 October 2020}, pages 5036--5040. {ISCA}.

\bibitem[{Horlock and King(2003)}]{Horlock2003DiscriminativeMF}
James Horlock and Simon King. 2003.
\newblock \href {https://doi.org/10.21437/Eurospeech.2003-737} {{Discriminative
  methods for improving named entity extraction on speech data}}.
\newblock In \emph{Proc. 8th European Conference on Speech Communication and
  Technology (Eurospeech 2003)}, pages 2765--2768.

\bibitem[{Jang et~al.(2017)Jang, Gu, and Poole}]{jang2016categorical}
Eric Jang, Shixiang Gu, and Ben Poole. 2017.
\newblock \href {https://openreview.net/forum?id=rkE3y85ee} {Categorical
  reparameterization with gumbel-softmax}.
\newblock In \emph{5th International Conference on Learning Representations,
  {ICLR} 2017, Toulon, France, April 24-26, 2017, Conference Track
  Proceedings}. OpenReview.net.

\bibitem[{Jurafsky and Martin(2009)}]{Jurafsky}
Dan Jurafsky and James~H. Martin. 2009.
\newblock \href {https://www.worldcat.org/oclc/315913020} {\emph{Speech and
  language processing: an introduction to natural language processing,
  computational linguistics, and speech recognition, 2nd Edition}}.
\newblock Prentice Hall series in artificial intelligence. Prentice Hall,
  Pearson Education International.

\bibitem[{Kubala et~al.(1998)Kubala, Schwartz, Stone, and
  Weischedel}]{kubala1998named}
Francis Kubala, Richard Schwartz, Rebecca Stone, and Ralph Weischedel. 1998.
\newblock \href
  {https://www.researchgate.net/profile/Richard-Schwartz-2/publication/2347872_Named_Entity_Extraction_From_Speech/links/54d250700cf28e069723d9d6/Named-Entity-Extraction-From-Speech.pdf}
  {Named entity extraction from speech}.
\newblock In \emph{Proceedings of DARPA Broadcast News Transcription and
  Understanding Workshop}, pages 287--292. Citeseer.

\bibitem[{Kudo and Richardson(2018)}]{kudo2018sentencepiece}
Taku Kudo and John Richardson. 2018.
\newblock \href {https://doi.org/10.18653/v1/d18-2012} {Sentencepiece: {A}
  simple and language independent subword tokenizer and detokenizer for neural
  text processing}.
\newblock In \emph{Proceedings of the 2018 Conference on Empirical Methods in
  Natural Language Processing, {EMNLP} 2018: System Demonstrations, Brussels,
  Belgium, October 31 - November 4, 2018}, pages 66--71. Association for
  Computational Linguistics.

\bibitem[{Lafferty et~al.(2001)Lafferty, McCallum, and Pereira}]{CRF_og}
John~D. Lafferty, Andrew McCallum, and Fernando C.~N. Pereira. 2001.
\newblock \href
  {https://repository.upenn.edu/cgi/viewcontent.cgi?article=1162&context=cis_papers}
  {Conditional random fields: Probabilistic models for segmenting and labeling
  sequence data}.
\newblock In \emph{Proceedings of the Eighteenth International Conference on
  Machine Learning}, ICML '01, page 282–289, San Francisco, CA, USA. Morgan
  Kaufmann Publishers Inc.

\bibitem[{Lample et~al.(2016)Lample, Ballesteros, Subramanian, Kawakami, and
  Dyer}]{lample-etal-2016-neural}
Guillaume Lample, Miguel Ballesteros, Sandeep Subramanian, Kazuya Kawakami, and
  Chris Dyer. 2016.
\newblock \href {https://doi.org/10.18653/v1/N16-1030} {Neural architectures
  for named entity recognition}.
\newblock In \emph{Proceedings of the 2016 Conference of the North {A}merican
  Chapter of the Association for Computational Linguistics: Human Language
  Technologies}, pages 260--270, San Diego, California. Association for
  Computational Linguistics.

\bibitem[{Ma and Hovy(2016)}]{ma-hovy-2016-end}
Xuezhe Ma and Eduard Hovy. 2016.
\newblock \href {https://doi.org/10.18653/v1/P16-1101} {End-to-end sequence
  labeling via bi-directional {LSTM}-{CNN}s-{CRF}}.
\newblock In \emph{Proceedings of the 54th Annual Meeting of the Association
  for Computational Linguistics (Volume 1: Long Papers)}, pages 1064--1074,
  Berlin, Germany. Association for Computational Linguistics.

\bibitem[{McCallum et~al.(2000)McCallum, Freitag, and Pereira}]{memm}
Andrew McCallum, Dayne Freitag, and Fernando C.~N. Pereira. 2000.
\newblock \href {http://www.ai.mit.edu/courses/6.891-nlp/READINGS/maxent.pdf}
  {Maximum entropy markov models for information extraction and segmentation}.
\newblock In \emph{Proceedings of the Seventeenth International Conference on
  Machine Learning}, ICML '00, page 591–598, San Francisco, CA, USA. Morgan
  Kaufmann Publishers Inc.

\bibitem[{Mitchell et~al.(2018)Mitchell, Cohen, Hruschka, Talukdar, Yang,
  Betteridge, Carlson, Dalvi, Gardner, Kisiel, Krishnamurthy, Lao, Mazaitis,
  Mohamed, Nakashole, Platanios, Ritter, Samadi, Settles, Wang, Wijaya, Gupta,
  Chen, Saparov, Greaves, and Welling}]{NELL}
T.~Mitchell, W.~Cohen, E.~Hruschka, P.~Talukdar, B.~Yang, J.~Betteridge,
  A.~Carlson, B.~Dalvi, M.~Gardner, B.~Kisiel, J.~Krishnamurthy, N.~Lao,
  K.~Mazaitis, T.~Mohamed, N.~Nakashole, E.~Platanios, A.~Ritter, M.~Samadi,
  B.~Settles, R.~Wang, D.~Wijaya, A.~Gupta, X.~Chen, A.~Saparov, M.~Greaves,
  and J.~Welling. 2018.
\newblock \href {https://doi.org/10.1145/3191513} {Never-ending learning}.
\newblock \emph{Commun. ACM}, 61(5):103–115.

\bibitem[{Morwal et~al.(2012)Morwal, Jahan, and Chopra}]{HMM_article}
Sudha Morwal, Nusrat Jahan, and Deepti Chopra. 2012.
\newblock \href {https://doi.org/10.5121/ijnlc.2012.1402} {Named entity
  recognition using hidden markov model (hmm)}.
\newblock \emph{International Journal on Natural Language Computing}, 1:15--23.

\bibitem[{Nguyen and Yu(2021)}]{nguyen-yu-2021-improving}
Minh Nguyen and Zhou Yu. 2021.
\newblock \href {https://aclanthology.org/2021.sigdial-1.6} {Improving named
  entity recognition in spoken dialog systems by context and speech pattern
  modeling}.
\newblock In \emph{Proceedings of the 22nd Annual Meeting of the Special
  Interest Group on Discourse and Dialogue}, pages 45--55, Singapore and
  Online. Association for Computational Linguistics.

\bibitem[{Nystrom et~al.(2015)Nystrom, Levine, Roskies, and
  Scott}]{nystrom2015bridges}
Nicholas~A. Nystrom, Michael~J. Levine, Ralph~Z. Roskies, and J.~Ray Scott.
  2015.
\newblock \href {https://doi.org/10.1145/2792745.2792775} {Bridges: a uniquely
  flexible {HPC} resource for new communities and data analytics}.
\newblock In \emph{Proceedings of the 2015 {XSEDE} Conference: Scientific
  Advancements Enabled by Enhanced Cyberinfrastructure, St. Louis, MO, USA,
  July 26 - 30, 2015}, pages 30:1--30:8. {ACM}.

\bibitem[{Palmer and Ostendorf(2001)}]{palmer-ostendorf-2001-improving}
David~D. Palmer and Mari Ostendorf. 2001.
\newblock \href {https://aclanthology.org/H01-1034} {Improving information
  extraction by modeling errors in speech recognizer output}.
\newblock In \emph{Proceedings of the First International Conference on Human
  Language Technology Research}.

\bibitem[{Parada et~al.(2011)Parada, Dredze, and Jelinek}]{parada2011oov}
Carolina Parada, Mark Dredze, and Frederick Jelinek. 2011.
\newblock \href
  {http://www.isca-speech.org/archive/interspeech\_2011/i11\_2085.html} {{OOV}
  sensitive named-entity recognition in speech}.
\newblock In \emph{{INTERSPEECH} 2011, 12th Annual Conference of the
  International Speech Communication Association, Florence, Italy, August
  27-31, 2011}, pages 2085--2088. {ISCA}.

\bibitem[{Park et~al.(2019)Park, Chan, Zhang, Chiu, Zoph, Cubuk, and
  Le}]{specaugment}
Daniel~S. Park, William Chan, Yu~Zhang, Chung{-}Cheng Chiu, Barret Zoph,
  Ekin~D. Cubuk, and Quoc~V. Le. 2019.
\newblock \href {https://doi.org/10.21437/Interspeech.2019-2680} {Specaugment:
  {A} simple data augmentation method for automatic speech recognition}.
\newblock In \emph{Interspeech 2019, 20th Annual Conference of the
  International Speech Communication Association, Graz, Austria, 15-19
  September 2019}, pages 2613--2617. {ISCA}.

\bibitem[{Paszke et~al.(2019)Paszke, Gross, Massa, Lerer, Bradbury, Chanan,
  Killeen, Lin, Gimelshein, Antiga, Desmaison, K{\"{o}}pf, Yang, DeVito,
  Raison, Tejani, Chilamkurthy, Steiner, Fang, Bai, and Chintala}]{pytorch}
Adam Paszke, Sam Gross, Francisco Massa, Adam Lerer, James Bradbury, Gregory
  Chanan, Trevor Killeen, Zeming Lin, Natalia Gimelshein, Luca Antiga, Alban
  Desmaison, Andreas K{\"{o}}pf, Edward~Z. Yang, Zachary DeVito, Martin Raison,
  Alykhan Tejani, Sasank Chilamkurthy, Benoit Steiner, Lu~Fang, Junjie Bai, and
  Soumith Chintala. 2019.
\newblock \href
  {https://proceedings.neurips.cc/paper/2019/hash/bdbca288fee7f92f2bfa9f7012727740-Abstract.html}
  {Pytorch: An imperative style, high-performance deep learning library}.
\newblock In \emph{Advances in Neural Information Processing Systems 32: Annual
  Conference on Neural Information Processing Systems 2019, NeurIPS 2019,
  December 8-14, 2019, Vancouver, BC, Canada}, pages 8024--8035.

\bibitem[{Povey et~al.(2011)Povey, Ghoshal, Boulianne, Burget, Glembek, Goel,
  Hannemann, Motlicek, Qian, Schwarz, Silovsky, Stemmer, and
  Vesely}]{povey2011kaldi}
Daniel Povey, Arnab Ghoshal, Gilles Boulianne, Lukas Burget, Ondrej Glembek,
  Nagendra Goel, Mirko Hannemann, Petr Motlicek, Yanmin Qian, Petr Schwarz, Jan
  Silovsky, Georg Stemmer, and Karel Vesely. 2011.
\newblock \href
  {https://infoscience.epfl.ch/record/192584/files/Povey_ASRU2011_2011.pdf}
  {The {K}aldi speech recognition toolkit}.
\newblock In \emph{IEEE 2011 Workshop on Automatic Speech Recognition and
  Understanding}. IEEE Signal Processing Society.

\bibitem[{Ramshaw and Marcus(1995)}]{ramshaw1999text}
Lance~A. Ramshaw and Mitch Marcus. 1995.
\newblock \href {https://aclanthology.org/W95-0107/} {Text chunking using
  transformation-based learning}.
\newblock In \emph{Third Workshop on Very Large Corpora, VLC@ACL 1995,
  Cambridge, Massachusetts, USA, June 30, 1995}.

\bibitem[{Rao et~al.(2020)Rao, Raju, Dheram, Bui, and Rastrow}]{raoetal2020}
Milind Rao, Anirudh Raju, Pranav Dheram, Bach Bui, and Ariya Rastrow. 2020.
\newblock \href {https://doi.org/10.21437/Interspeech.2020-2976} {Speech to
  semantics: Improve {ASR} and {NLU} jointly via all-neural interfaces}.
\newblock In \emph{Interspeech 2020, 21st Annual Conference of the
  International Speech Communication Association, Virtual Event, Shanghai,
  China, 25-29 October 2020}, pages 876--880. {ISCA}.

\bibitem[{Saxon et~al.(2021)Saxon, Choudhary, McKenna, and
  Mouchtaris}]{SaxonCMM21}
Michael Saxon, Samridhi Choudhary, Joseph~P. McKenna, and Athanasios
  Mouchtaris. 2021.
\newblock \href {https://doi.org/10.21437/Interspeech.2021-1826} {End-to-end
  spoken language understanding for generalized voice assistants}.
\newblock In \emph{Interspeech 2021, 22nd Annual Conference of the
  International Speech Communication Association, Brno, Czechia, 30 August - 3
  September 2021}, pages 4738--4742. {ISCA}.

\bibitem[{Shon et~al.(2022)Shon, Pasad, Wu, Brusco, Artzi, Livescu, and
  Han}]{SLUE}
Suwon Shon, Ankita Pasad, Felix Wu, Pablo Brusco, Yoav Artzi, Karen Livescu,
  and Kyu~J. Han. 2022.
\newblock \href {https://doi.org/10.1109/ICASSP43922.2022.9746137} {{SLUE:} new
  benchmark tasks for spoken language understanding evaluation on natural
  speech}.
\newblock In \emph{{IEEE} International Conference on Acoustics, Speech and
  Signal Processing, {ICASSP} 2022, Virtual and Singapore, 23-27 May 2022},
  pages 7927--7931. {IEEE}.

\bibitem[{Towns et~al.(2014)Towns, Cockerill, Dahan, Foster, Gaither, Grimshaw,
  Hazlewood, Lathrop, Lifka, Peterson, Roskies, Scott, and
  Wilkins{-}Diehr}]{xsede}
John Towns, Timothy Cockerill, Maytal Dahan, Ian~T. Foster, Kelly~P. Gaither,
  Andrew~S. Grimshaw, Victor Hazlewood, Scott~A. Lathrop, David Lifka,
  Gregory~D. Peterson, Ralph Roskies, J.~Ray Scott, and Nancy Wilkins{-}Diehr.
  2014.
\newblock \href {https://doi.org/10.1109/MCSE.2014.80} {{XSEDE:} accelerating
  scientific discovery}.
\newblock \emph{Comput. Sci. Eng.}, 16(5):62--74.

\bibitem[{Tran et~al.(2018)Tran, Toshniwal, Bansal, Gimpel, Livescu, and
  Ostendorf}]{Tran_SLU}
Trang Tran, Shubham Toshniwal, Mohit Bansal, Kevin Gimpel, Karen Livescu, and
  Mari Ostendorf. 2018.
\newblock \href {https://doi.org/10.18653/v1/n18-1007} {Parsing speech: a
  neural approach to integrating lexical and acoustic-prosodic information}.
\newblock In \emph{Proceedings of the 2018 Conference of the North American
  Chapter of the Association for Computational Linguistics: Human Language
  Technologies, {NAACL-HLT} 2018, New Orleans, Louisiana, USA, June 1-6, 2018,
  Volume 1 (Long Papers)}, pages 69--81. Association for Computational
  Linguistics.

\bibitem[{Vaswani et~al.(2017)Vaswani, Shazeer, Parmar, Uszkoreit, Jones,
  Gomez, Kaiser, and Polosukhin}]{vasvani}
Ashish Vaswani, Noam Shazeer, Niki Parmar, Jakob Uszkoreit, Llion Jones,
  Aidan~N. Gomez, Lukasz Kaiser, and Illia Polosukhin. 2017.
\newblock \href
  {https://proceedings.neurips.cc/paper/2017/hash/3f5ee243547dee91fbd053c1c4a845aa-Abstract.html}
  {Attention is all you need}.
\newblock In \emph{Advances in Neural Information Processing Systems 30: Annual
  Conference on Neural Information Processing Systems 2017, December 4-9, 2017,
  Long Beach, CA, {USA}}, pages 5998--6008.

\bibitem[{Watanabe et~al.(2018)Watanabe, Hori, Karita, Hayashi, Nishitoba,
  Unno, Soplin, Heymann, Wiesner, Chen, Renduchintala, and Ochiai}]{espnet}
Shinji Watanabe, Takaaki Hori, Shigeki Karita, Tomoki Hayashi, Jiro Nishitoba,
  Yuya Unno, Nelson Enrique~Yalta Soplin, Jahn Heymann, Matthew Wiesner, Nanxin
  Chen, Adithya Renduchintala, and Tsubasa Ochiai. 2018.
\newblock \href {https://doi.org/10.21437/Interspeech.2018-1456} {Espnet:
  End-to-end speech processing toolkit}.
\newblock In \emph{Interspeech 2018, 19th Annual Conference of the
  International Speech Communication Association, Hyderabad, India, 2-6
  September 2018}, pages 2207--2211. {ISCA}.

\bibitem[{Yan et~al.(2019)Yan, Deng, Li, and Qiu}]{transformer-crf}
Hang Yan, Bocao Deng, Xiaonan Li, and Xipeng Qiu. 2019.
\newblock \href {https://doi.org/10.48550/ARXIV.1911.04474} {Tener: Adapting
  transformer encoder for named entity recognition}.

\bibitem[{Yu et~al.(2019)Yu, Cohn, Yang, Chen, Wen, Zhang, Zhou, Jesse, Chau,
  Bhowmick, Iyer, Sreenivasulu, Davidson, Bhandare, and Yu}]{socialbot}
Dian Yu, Michelle Cohn, Yi~Mang Yang, Chun{-}Yen Chen, Weiming Wen, Jiaping
  Zhang, Mingyang Zhou, Kevin Jesse, Austin Chau, Antara Bhowmick, Shreenath
  Iyer, Giritheja Sreenivasulu, Sam Davidson, Ashwin Bhandare, and Zhou Yu.
  2019.
\newblock \href {https://doi.org/10.18653/v1/D19-3014} {Gunrock: {A} social bot
  for complex and engaging long conversations}.
\newblock In \emph{Proceedings of the 2019 Conference on Empirical Methods in
  Natural Language Processing and the 9th International Joint Conference on
  Natural Language Processing, {EMNLP-IJCNLP} 2019, Hong Kong, China, November
  3-7, 2019 - System Demonstrations}. Association for Computational
  Linguistics.

\bibitem[{Zhai et~al.(2004)Zhai, Fung, Schwartz, Carpuat, and
  Wu}]{zhai2004using}
Lu{-}Feng Zhai, Pascale Fung, Richard~M. Schwartz, Marine Carpuat, and Dekai
  Wu. 2004.
\newblock \href {https://aclanthology.org/N04-4010/} {Using n-best lists for
  named entity recognition from chinese speech}.
\newblock In \emph{Proceedings of {HLT-NAACL} 2004: Short Papers, Boston,
  Massachusetts, USA, May 2-7, 2004}. The Association for Computational
  Linguistics.

\bibitem[{Zhou et~al.(2015)Zhou, Suominen, and Hanlen}]{cascaded2}
Liyuan Zhou, Hanna Suominen, and Leif Hanlen. 2015.
\newblock \href {https://doi.org/10.1145/2802558.2814646} {Evaluation data and
  benchmarks for cascaded speech recognition and entity extraction}.
\newblock In \emph{Proceedings of the Third Edition Workshop on Speech,
  Language \& Audio in Multimedia}, SLAM '15, page 15–18, New York, NY, USA.
  Association for Computing Machinery.

\end{thebibliography}
\bibliographystyle{acl_natbib}

\appendix

\section{Appendix}

\subsection{Applications of SLU}
SLU is an essential component of many commercial devices like voice assistants, home assistants \cite{socialbot, snips-voice-platform} and spoken dialog systems \cite{nguyen-yu-2021-improving} that map speech to executable commands on a daily basis. One of the key applications of SLU is to extract key mentions like entities from a user command to take appropriate actions. As a result, several datasets~\cite{SLURP,SLUE,earnings21} have been proposed to build understanding systems for spoken utterances. 

\subsection{Dataset Description}
\label{sec:data}

We evaluated our proposed approach on publicly available SLU datasets, namely SLUE~\cite{SLUE} and SLURP~\cite{SLURP} datasets on the task of Named Entity Recognition (NER) from naturally available speech. SLURP is a linguistically diverse and challenging spoken language understanding benchmark that consists of single-turn user conversation with a home assistant, annotated with both intent and entities. Similar to the approach followed in our prior work \cite{SLURP,ESPnet-SLU}, we bootstrap our train set with 43 hours of synthetic data for all our experiments. We evaluate our approach using SLU-F1 \cite{SLURP}, a metric for spoken entity prediction, and Label F1, which considers only entity-tag predictions.

SLUE is a recently released SLU benchmark that focuses on Spoken Language Understanding from limited labeled training data. Specifically, it consists of SLUE VoxPopuli dataset that can be used for building systems for ASR and NER. Similar to \cite{SLUE}, we evaluate our systems using two micro-averaged F1 scores, the first score that evaluates both named entity and tag pairs is referred to as F1, and the second that evaluates only entity-tag phrases is referred to as Label-F1. Note that the released test sets are blind without ground truth labels, and hence we compare different methods using the development set.

The dataset download and evaluation links for SLURP can be found here - \url{https://github.com/pswietojanski/slurp} and for SLUE here - \url{https://github.com/asappresearch/slue-toolkit}. The datasets have been processed and prepared using ESPnet, SLURP - \url{https://github.com/espnet/espnet/tree/master/egs2/slurp_entity} and SLUE - \url{https://github.com/espnet/espnet/tree/master/egs2/slue-voxpopuli}

\begin{table}[t]
  \caption{Overview of the two publicly available SLU datasets~\cite{SLUE,SLURP} used for our experiments.}
  \label{tab:slue-datasets}
  \centering
  \resizebox{0.95\linewidth}{!}{
  \begin{tabular}{lccc}
    \toprule
    \multirow{2}{*}{Dataset} & \multicolumn{3}{c}{Size (utterances / hours)}\\\cmidrule(lr){2-4}
    & Train & Dev & Test\\
    \midrule
    SLURP & 11,514 / 40.2 & 2,033 / 6.9 & 2,974 /10.3\\
    SLUE-VoxPopuli & 5,000 / 14.5 & 1,753 / 5.0 & 1,842 / 4.9\\
    \bottomrule
  \end{tabular}
  }
\end{table}

\subsection{Experimental Setup}
\label{sec:exp_setup}

Our models are implemented in PyTorch~\cite{pytorch}, and the experiments are conducted using the ESPnet-SLU toolkit~\cite{ESPnet-SLU}. 

\subsubsection{Speech Preprocessing}
Speech inputs are globally mean-variance normalized 80 dimensional logmel filterbanks using a 16kHz sampling and window of 512 frames and a 128 hop length. We apply speed perturbation for the under-resourced dataset of SLUE of 0.9 and 1.1 to increase the samples. We also apply specaugmentation \cite{specaugment} on both datasets. We also remove all examples smaller than 0.1 seconds and larger than 20 seconds from the training data.

\subsubsection{Text Processing}
For the cascaded system, we process ASR transcripts $S$ using bpe tokenization \cite{kudo2018sentencepiece} and train ASR models to generate bpe subtokens. We use bpe size of 500 for SLURP and 1000 for SLUE dataset. For the direct E2E models, we predict the enriched label sequence $Y^{e}$ using the same bpe size as the ASR models in cascaded sequence. Similarly, compositional models also use the same bpe size to generate the ASR transcripts.

For creating the BIO tags we modify the data preparation such that we take the entities for each utterance and create a ``label utterance''. This consists of one-to-one mapping of the label tags with the words and Begin (B), Inside (I) and Outside (O) marked for each label. After performing BPE tokenization we add $\varnothing$ for every subtoken of the word. We have attached the data preparation code.

\subsubsection{Model and Training Hyperparameters}
We run parameter search for both direct end-to-end and our compositional end-to-end systems using the same model search space (\Tref{tab:hp-st-tr}). In this section, we will describe our best architecture for both direct and compositional E2E systems.

\paragraph{Direct E2E SLU systems}
 After searching through hyperparameter space, our Direct E2E SLU systems consists of 12-layer Conformer~\cite{conformer} encoder and a 6-layer Transformer~\cite{vasvani} decoder with 8 attention heads for SLURP dataset. We use a dropout of 0.1, output dim of 512 and feedforward dim of 2048, giving a total parameter size of 109.3 M.

For SLUE dataset,  we found 12-layer Conformer with 4 attention heads and decoder is a 6-layer Transformer with 4 attention heads to give best validation performance. We use a dropout of 0.1, output dim of 256 and feedforward dim of 1024 in encoder and 2048 in the decoder, giving a total parameter size of 31.2 M. 

\paragraph{Compositional E2E SLU systems}
Our Compositional model which uses Direct E2E SLU formulation consists of 12-layer conformer block for encoder, 6-layer transformer block for decoder in it's ASR component and 4-layer transformer encoder and 6-layer transformer decoder in it's NLU component. Each of these attention blocks consist of 8 attention heads, dropout of 0.1, output dim of 512, feedforward dim of 2048, giving a total of 153.9M parameters in SLURP dataset. For SLUE dataset, each attention block has 4 attention heads, dropout of 0.1, output dim of 256, feedforward dimension of 1024 in encoder and 2048 in decoder, giving a total parameter size of 46.8M.

Our Composition model with Proposed NLU formulation replaces NLU component in Direct E2E formulation with 8-layer transformer encoder followed by linear layer. All these attention blocks consist of 8 attention heads, dropout of 0.1, output dim of 512, feedforward dim of 2048, giving a total of 142.9M parameters in SLURP dataset.  For SLUE dataset, each of these attention blocks have 4 attention heads, dropout of 0.1, output dim of 256, feedforward dimension of 1024 in encoder and 2048 in decoder, giving a total parameter size of 43.8M. Our NLU component can further attend to speech representations using cross attention~\cite{dalmia-etal-2021-searchable}. We further implement CRF loss using publicly available python library~\footnote{\url{https://pytorch-crf.readthedocs.io}}.

The loss from the ASR ($\mathcal{L}^\text{asr}$) and NLU ($\mathcal{L}^\text{nlu}$) subnet are combined combined as follows 
\begin{equation*}
    \mathcal{L} = \mathcal{L}^\text{asr} + \alpha\mathcal{L}^\text{nlu}
\end{equation*}
We search alpha values over [0.3, 0.4, 0.5, 0.6] and found 0.6 to be best for SLURP and 0.3 for SLUE.

\begin{table}[t]
  \centering
  \resizebox {0.8\linewidth} {!} {
  \begin{tabular}{lr}
    \toprule
    Hyperparameter & Value \\
    \midrule
    Output Size & [256, 512] \\
    Attention Heads & [4, 8] \\
    Number of blocks & [4, 6, 8, 12] \\
    Hidden Dropout & [0.1, 0.2] \\
    Attention dropout & [0.1, 0.2] \\
    Position dropout & [0.1, 0.2] \\
    Activation dropout & [0.1, 0.2] \\
    Src Activation dropout & [0.1, 0.2] \\
    Batch size & [ 50, 64]\\
    LR schedule & [inv. sqrt., exp. lr.]\\
    Max learning rate & [0.001, 0.002, 0.003] \\
    Warmup steps & [5000, 15000, 25000] \\
    Number of steps & [50, 70, 100] \\
    Adam eps  & 1e-9 \\
    Adam betas  & (0.9, 0.98)\\
    Weight decay & 0.000001\\
    \bottomrule
  \end{tabular}
  }
  \caption{Model and Training Search for SLU Models.}
  \label{tab:hp-st-tr}
\end{table}

\begin{table}[ht]
\resizebox {\linewidth} {!} {
  \centering
\begin{tabular}{lcc|cc}
\toprule
& \multicolumn{2}{c|}{SLURP} & \multicolumn{2}{c}{SLUE} \\ 
\cmidrule(r){2-5}
Model & SLU F1 & Label F1 & F1 & Label F1 \\ 
 \midrule
Casacaded SLU (Ours) & 76.9 & 83.9 & 48.6 & 63.9  \\
Direct E2E SLU (Ours) &  79.2 & 85.4 & 54.7 & 67.6 \\  %
\midrule
Compositional E2E SLU \\
w/ Direct E2E formulation (\Sref{sec:slu}) & 79.3 & 86.6 & 50.0 & 68.0\\ %
w/ Proposed NLU formulation (\Sref{sec:compositional_e2e}) \\ 
\hphantom{000}CRF w/ Speech Attention (SA) & 79.9 & 87.0  & 59.4 & 73.6\\ %
\hphantom{000}Token Classification w/ SA & 79.8 & 86.9 & 60.3 & 73.7\\  %
\hphantom{000}\hphantom{000}w/o Speech Attention & 79.7 & 87.0 & 59.0 & 73.6 \\  %
\bottomrule
\end{tabular}
}
\caption{Results presenting the micro F1 performance for all models using CRF and Token Classification modeling on development set for SLURP and SLUE} 
\label{tab:devel-results}
\end{table}

\subsubsection{Decoding Hyperparameters}
We keep the same decoding parameter of beam size and penalty as that of \citet{ESPnet-SLU}. For direct E2E systems and our models CTC weight of 0.1 worked best. We searched over CTC weight of [0, 0.1, 0.3, 0.5]. 

\subsubsection{Development Results}
We use F1 scores on the validation data to select the best hyperparameters. \Tref{tab:devel-results} presents the validation performances for our models. 

\begin{figure*}[t!]
    \centering
    \includegraphics[width=\linewidth]{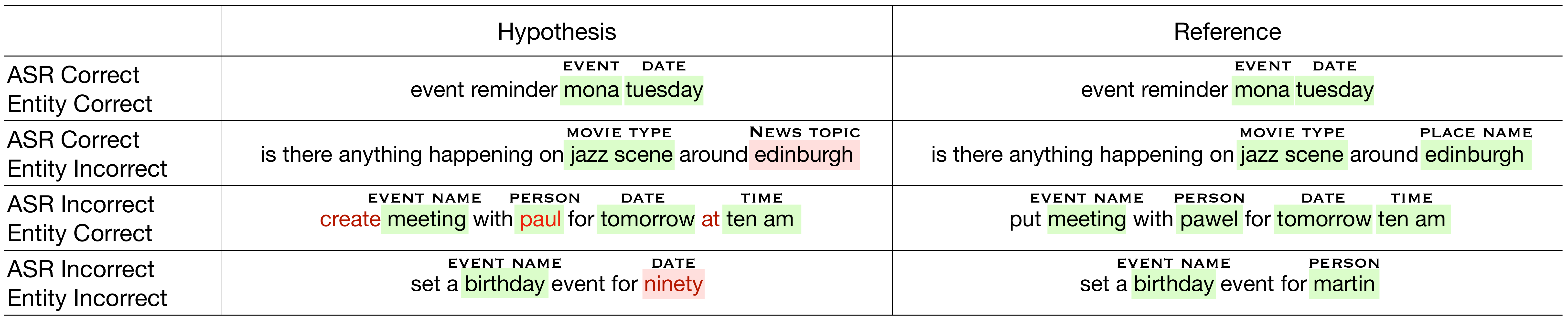}
    \caption{Qualitative examples of our compositional E2E SLU model for various error categories. We can observe that in the first case, the model is correctly able to predict both entity types and mentions even when the name ``mona'' is not a common name for an event. In the second case, even though it predicts the correct ASR transcript, it mislabels ``Edinburgh'' as a news topic since the phrase ``is there anything happening'' usually occurs with news topics. In the third case, even though it makes a mistake in the person name, the model correctly tags it as a person. Finally, the model incorrectly generates the word ``ninety,'' and this error gets propagated to the NLU component through token representations which then predicts entity type ``date''. This analysis shows that the alignment between ASR and NLU outputs can help us gain better insights into model performance.}
    \label{fig:qualitative}
\end{figure*}
\subsubsection{Compute Infrastructure}
Our models were trained using mixed precision training on either a100, v100 or A6000 on our compute infrastructure depending on their availability. Depending on the GPU and the file i/o latency, the training time ranged from 4-7 hours for SLUE, while for SLURP the training time ranged from 12-18 hours.

\subsubsection{External ASR and NLU components}
For the experiments in ~\Tref{tab:pre-trained}, we used ASR and NLU models trained on external data. For the ASR fine-tuning we used an ESPnet model ~\footnote{\url{https://zenodo.org/record/4630406}} trained on the GigaSpeech dataset~\cite{gigaspeech}. This model has the same architecture as the baseline direct E2E model on SLURP. We initialize both the encoder and decoder for direct E2E SLU and the ASR sub-net for the compositional E2E SLU model. For NLU fine-tuning we used Canine ~\cite{canine}, a character based BERT language model, which exhibits strong performance on named entity recognition while being able to model token sizes comparable to our SLU systems.~\footnote{\url{https://huggingface.co/google/canine-s}} We initialize our NLU sub-network without speech attention with Canine and keep the model parameters fixed during training. For finding the best parameters we only tuned the learning rate and LR schedule from \Tref{tab:hp-st-tr} and report the best numbers among CRF and Token Classification loss.

For using External ASR Transcripts, we trained an ASR system initialized using GigaSpeech and WavLM~\cite{WavLM} respectively. They were then fine-tuned on the respective datasets. These systems achieve 10.0\% WER and 9.2\% WER on SLURP and SLUE respectively.

\subsection{Error Categorization}
\definecolor{cfred}{HTML}{e9a3c9}
\definecolor{cfgreen}{HTML}{a1d76a}

\begin{table}[t!]
  \centering
\resizebox {\linewidth} {!} {
\begin{tabular}{l|cc|cc}
\toprule
& \multicolumn{2}{c|}{Entity Correct} & \multicolumn{2}{|c}{Entity Incorrect} \\
\cmidrule{2-3}\cmidrule{4-5}
& Model & \# Examples & Model & \# Examples \\
 \midrule
 \multirow{2}{*}{ASR Correct} & \cellcolor{cfgreen} w/ SA & \cellcolor{cfgreen} 8520 & \cellcolor{cfred} w/ SA &  \cellcolor{cfred} 465 \\
 & \cellcolor{cfgreen} w/o SA & \cellcolor{cfgreen} 8501 &  \cellcolor{cfred} w/o SA &  \cellcolor{cfred} 474  \\
 \midrule
 \multirow{2}{*}{ASR Incorrect} &  \cellcolor{cfred} w/ SA &  \cellcolor{cfred} 1568 &  \cellcolor{cfred} w/ SA &  \cellcolor{cfred} 1343 \\
 &  \cellcolor{cfred} w/o SA &  \cellcolor{cfred} 1585 &  \cellcolor{cfred} w/o SA &  \cellcolor{cfred} 1336 \\
\bottomrule
\end{tabular}
}
\caption{Number of examples per error category of our compositional E2E SLU systems with/without Speech Attention on SLURP test set. There are four categories depending on whether mistakes are made by ASR or NLU component. Note that the \colorbox{cfgreen}{first quadrant} lists \# of \colorbox{cfgreen}{\textit{correct}} examples, while the \colorbox{cfred}{rest} list \colorbox{cfred}{\textit{incorrect}} ones. Direct E2E systems cannot offer such categorizations particularly for incorrect entities as there is no alignment between ASR and NLU outputs.} 
\label{tab:error-cat}
\vspace{-1.0em}
\end{table}

\label{sec:error_cat}
The predictions made by our compositional E2E SLU model can be categorized into different buckets on the basis of the errors by ASR or NER component. \Tref{tab:error-cat} demonstrates this behavior by categorizing the errors of our compositional E2E model trained with and without speech attention. Most of the performance differences between compositional E2E SLU model w/ and w/o speech attention are caused by the kinds of errors where the ASR predictions are inaccurate, but the NLU module is nevertheless able to recover the correct entity type from the utterance. This confirms our intuition that cross attention on speech representations can help the NLU module to recover from mistakes made during ``recognizing'' spoken mentions. We also present anecdotes for each of these error categories in \Fref{fig:qualitative}. This further emphasizes the transparency in our compositional E2E SLU models. Due to the lack of one-to-one alignment between ASR and Sequence Labeling, such analysis is not possible in direct E2E SLU systems, making it particularly difficult to categorize errors when the entity prediction is wrong.

\end{document}